\pdfoutput=1

\documentclass[11pt]{article}

\usepackage[]{acl}
\usepackage{times}
\usepackage{latexsym}

\usepackage{algorithm}
\usepackage{algpseudocode}
\usepackage{amsmath}
\usepackage{array}

\usepackage{booktabs}

\usepackage[T1]{fontenc}

\usepackage[utf8]{inputenc}

\usepackage{microtype}

\usepackage{inconsolata}

\usepackage{pgfplots}
\pgfplotsset{compat=1.18} 
\usepackage{pgf-pie}
\usepackage{tabularx}
\usepackage{adjustbox}
\usetikzlibrary{patterns}

\usepackage{hyperref}

\title{Archimedes-AUEB at SemEval-2024 Task 5: LLM explains Civil Procedure}

\author{ Odysseas S. Chlapanis$^{1, 3}$, \hspace{0.5mm} Ion Androutsopoulos$^{1, 3}$ \and Dimitrios Galanis$^{2, 3}$\\
$^{1}$Department of Informatics, Athens University of Economics and Business, Greece\\
$^{2}$Institute for Language and Speech Processing,  Athena Research Center, Greece\\$^{3}$Archimedes Unit, Athena Research Center, Greece\\ 
\texttt{odyhlapanis@aueb.gr, ion@aueb.gr, galanisd@athenarc.gr}
}

\begin{document}
\maketitle
\begin{abstract}
The SemEval task on Argument Reasoning in Civil Procedure is challenging in that it requires understanding legal concepts and inferring complex arguments.
Currently, most Large Language Models (LLM) excelling in the legal realm are principally purposed for classification tasks, hence their reasoning rationale is subject to contention.
The approach we advocate involves using a powerful teacher-LLM (ChatGPT) to extend the training dataset with explanations and generate synthetic data. The resulting data are then leveraged to fine-tune a small student-LLM.
Contrary to previous work, our explanations are not directly derived from the teacher's internal knowledge. Instead they are grounded in authentic human analyses, therefore delivering a superior reasoning signal.
Additionally, a new `mutation' method generates artificial data instances inspired from existing ones. 
We are publicly releasing the explanations as an extension to the original dataset, along with the synthetic dataset and the prompts that were used to generate both.
Our system ranked $15^\text{th}$ in the SemEval competition. It outperforms its own teacher and can produce explanations aligned with the original human analyses, as verified by legal experts. 
\end{abstract}
\section{Introduction}

\begin{table}[h]
\small
\begin{adjustbox}{width=0.5\textwidth,center}
\begin{tabularx}{\linewidth}{>{\raggedright\arraybackslash}X}
\toprule
{\centering \textbf{Introduction}} \\ 
\midrule
...\\
Klaxon: a federal diversity court should use the choice-of-law rules of the state in which it sits.\\
...\\
\midrule
{\centering \textbf{Question}}  \\
\midrule
A Rhode Island citizen goes skiing in Vermont. He falls and gets injured.
He brings a diversity action against the operator in federal court in Rhode Island.\\
...\\
The judge would probably apply\\
\midrule
{\centering \textbf{Answer (label: \textit{correct})}} \\
\midrule
Vermont law, because the accident happened there. \\
\midrule
\textbf{Expert Analysis}\\
\midrule
Rhode Island's choice-of-law rule calls for application of the law of the place of injury.\\
...\\
\midrule
\midrule
{\centering \textbf{Model Explanation}} \\
\midrule
\midrule
The federal district judge would probably apply Vermont law, as per Klaxon holding that the judge should use the local state's choice-of-law rules. \\
\bottomrule
\end{tabularx}
\end{adjustbox}
\caption{Example of a training instance and a system-generated explanation aligned with the expert's analysis.}
\label{tab:correct-example}
\end{table}

SemEval-2024 Task 5 \cite{task-5} concerns Legal Reasoning within the realm of US Civil Procedure, based on the \textit{Legal Argument Reasoning Task in Civil Procedure} \cite{legal_arg_reasoning_US_civil_procedure} dataset. 
It requires understanding legal concepts and advanced reasoning capabilities, such as the ability to grasp analogy-based arguments and identify contradictions. It is cast as a binary classification problem where, given a question and a candidate answer, the system has to respond whether the  candidate answer is correct or not. The questions and answers originate from a textbook widely adopted by US law schools.

Our system leverages two data augmentation strategies to enrich the provided training set.
The first strategy involves extending each data point with 
Chain-of-Thought \cite{wei2023chainofthought} style explanations originating from the author's original analysis and refined by GPT-3.5.
The second strategy generates additional synthetic examples inspired by the original training examples. The synthetic examples are also accompanied by explanations. We call this method \emph{data mutation}. The data generated from GPT-3.5 for both methods and the prompts used are publicly available.\footnote{\href{https://github.com/nlpaueb/multiple-choice-mutation}{github.com/nlpaueb/multiple-choice-mutation}}

We fine-tune a Llama-2 \cite{llama_2} model using both synthetic and original data, incorporating explanations, to develop a model capable of generating responses and explaining the reasoning supporting them. 
Our findings indicate that both data augmentation methods significantly improve performance.
We conducted ablation studies to analyze the effects of each method.
In addition, legal experts evaluate our model's explanations manually to offer us valuable insights about the quality of the produced explanations and the root causes of its errors.
\section{Background}

\subsection{Task Setup}
The dataset used in SemEval-2024 Task 5 is derived from a textbook on US civil procedure \cite{glannon}.
Each data instance consists of three parts: introduction, question, and answer (with the label of the answer), as shown in Table \ref{tab:correct-example}. The training and development instances also have an additional section, containing an analysis of the textbook's author (see `expert analysis' in Table~\ref{tab:correct-example}). The goal is to predict the label (correct or incorrect) of the given answer. The latter has the form of a completion of an incomplete text (the `question' part of the instance). 
Each chapter of the book addresses a specific topic, which is introduced in the `introduction'  of each instance. 

\subsection{Related Work}

In the last few years, several models were proposed for legal tasks. LegalBERT \cite{chalkidis-etal-2020-legal}, a BERT-based model \cite{devlin-etal-2019-bert} that has been further pre-trained on legal data, achieved state-of-the-art results in three downstream tasks. However, LegalBERT is unable to generate explanations for its decisions.
In subsequent work to address this limitation,
a new hierarchical extension of LegalBERT was proposed \cite{chalkidis-etal-2021-paragraph} that can select paragraphs of the input texts that justify its decisions, but it has to be trained with annotated data. Lawyer LLaMA \cite{lawyer-llama-report} enhanced the Chinese LLaMA \cite{chinesellama}, a descendant of LLaMA-1 \cite{llama_1}, with legal knowledge. This was achieved by supervised fine-tuning on a mix of synthetic and manually created legal Chinese datasets.

Chain-of-Thought (CoT) \cite{wei2023chainofthought} was introduced as  ``a series of intermediate natural language reasoning steps that lead to the final output''. Few-shot CoT prompting \cite{wei2023chainofthought} is a method that enhances the reasoning skills of LLMs by providing a few CoT demonstrations as exemplars in the prompt before asking the LLM for the final answer. 
Furthermore, it has been shown that CoT prompting can be improved by using a technique called self-consistency decoding \cite{wang2023selfconsistency}, where sampling is used during decoding to obtain multiple answers, and then the most consistent one is chosen (e.g., via majority voting).

Instruction tuning \cite{ouyang2022training} fine-tunes an LLM on a wide range of input requests and the desirable responses, in order to train the LLM to follow instructions.
WizardLM \cite{xu2023wizardlm} implements instruction tuning on synthetic instruction data of progressively increasing complexity. The synthetic instruction data are commonly generated by a powerful teacher-LLM.
Reinforced Self-Training (ReST) \cite{gulcehre2023reinforced} iteratively enhances synthetic data by automatically filtering out lower-quality samples and then utilizing the remaining synthetic data to 
improve the model that produces the synthetic data.
Orca \cite{mukherjee2023orca} introduces explanation tuning, which incorporates fine-tuning on CoT data generated by \mbox{GPT-4} to teach advanced reasoning skills to a smaller LLM. 

\section{System overview}
\label{sec:system}

\subsection{Method} 

Our approach is based on a small open-source LLM (LLama-2-7B), which is fine-tuned to generate CoT-like explanations supporting its predictions. A much larger teacher LLM (GPT-3.5) is utilized to acquire appropriate reasoning data (CoT explanations) with two data augmentation techniques. The first technique modifies the experts' analyses (Section \ref{hge}) to turn them into CoT explanations appropriate for fine-tuning. 
The second technique `mutates' an original training instance to generate a synthetic one that incorporates alternative fictitious elements, but requires the application of similar legal reasoning to arrive at the correct answer (Section \ref{mcm}).
Experiments show that both techniques boost the performance of the baseline model, which is the same LLM fine-tuned without explanations.

\subsection{Prompt structure}
To create appropriate reasoning data, GPT-3.5 has to perform challenging tasks and the key to achieve this is handcrafting clear and concise prompts. We follow prompt-engineering best practices \cite{bsharat2024principled} to create prompts for both data augmentation techniques. Each prompt has three parts: system instructions, a query-response pair and a new query that GPT-3.5 has to respond to as in the example. 
The system instructions are designed to a) delineate the model's role, typically attributed as a legal expert in US Civil Procedure, b) outline the task to be executed and c) specify the desired output format. 
Complete prompt examples are provided in Appendix \ref{sec:appendix_prompts}. 

\subsection{Human-guided explanations (HGE)}
\label{hge}
CoT explanations that will be leveraged for fine-tuning must adhere to a uniform and formal style.
Furthermore, we desire our model's explanations to be succinct so that reasoning fallacies can be easily detected by legal experts.
Although the expert analysis (Table~\ref{tab:correct-example}) provided by the dataset is similar to a CoT explanation, it is tailored for students and therefore more detailed and informal.
We refine the expert's analysis with the assistance of GPT-3.5 (Table~\ref{tab:hge} shows the prompt used) to align it with our desired properties.

\begin{table}[h!]
\begin{minipage}[t]{0.5\linewidth}
\small

\begin{tabularx}{\linewidth}{>{\raggedright\arraybackslash}X}
\toprule
\textbf{HGE}\\
\midrule
\textbf{Input:}\\
Training instance\\
\textbf{Prompt:}\\
Explain why the answer is correct/incorrect according to the analysis.\\
\textbf{ChatGPT's response:}\\
Explanation: ...\\
\bottomrule
\end{tabularx}
\end{minipage}%
\hfill
\begin{minipage}[t!]{0.45\linewidth}
\caption{The Human-Guided Explanations (HGE) prompt asks the LLM to provide a CoT explanation that aligns with the analysis of the legal expert and follows a specific format.}
\label{tab:hge}
\end{minipage}%
\end{table}

\subsection{Multiple choice mutation (MCM)}
\label{mcm}
Our goal with this technique is to generate artificial data that demand similar reasoning skills and legal knowledge to the original ones. You can see an example in Table \ref{tab:mutation} for the training instance in Table \ref{tab:correct-example}. We use GPT-3.5 as the teacher-LLM. 
We implement this in two different prompting stages.

\begin{table}[h]
\small
\begin{adjustbox}{width=0.5\textwidth,center}
\begin{tabularx}{\linewidth}{>{\raggedright\arraybackslash}X}
\toprule
{\centering \textbf{MCM example}} \\ 
\midrule
{\centering \textbf{Concept:}} \\ 
Choice of Law in Civil Procedure\\
\textbf{Question Background:}\\
Alex, a Florida resident, buys a rare art piece from Carter, a California resident.
The contract stipulates that any disputes arising from the agreement will be resolved in Arizona.
A disagreement arises over the authenticity of the artwork.
Alex sues Carter in federal court in Florida.\\
...\\
\textbf{Multiple Choice Question:}\\
Which state's contract law would most likely be applied in Alex's case against Carter?\\
{\centering \textbf{Options}:}\\
A) California (label: \textit{incorrect})\\
Explanation:\\ 
Carter's residence is not a determining factor as the contract specifies the choice of law.\\
B) Arizona (label: \textit{correct})\\
Explanation:\\
The contract explicitly states that any disputes will be resolved in Arizona.\\
C) Florida (label: \textit{incorrect})\\
Explanation: ...\\
D) New York (label: \textit{incorrect})\\
Explanation: ...\\
\bottomrule
\end{tabularx}
\end{adjustbox}
\caption{Example of the MCM algorithm applied on the training instance shown in Table~\ref{tab:correct-example}. Each option becomes a distinct mutated instance.}
\label{tab:mutation}
\end{table}

\begin{table}[h!]
\begin{minipage}[t]{0.45\linewidth}
\small

\begin{tabularx}{\linewidth}{>{\raggedright\arraybackslash}X}
\toprule
\textbf{MCM Stage A}\\
\midrule
\textbf{Input:}\\
Training instance\\
\textbf{Prompt:}\\
Generate a multiple choice question that illustrates the same concept with a different question background.\\
\textbf{ChatGPT's response:}\\
{Concept}\\ 
Mutated Question \\
(Question Background + Multiple Choice Question)\\
{Options: a, b, c, d}\\
Correct option: \textit{X}\\
\bottomrule
\end{tabularx}
\caption{In stage A of MCM, we generate a synthetic example question and four options, one of which is correct.}
\label{tab:mcm_a}
\end{minipage}%
\hfill
\begin{minipage}[t]{0.5\linewidth}
\small
\begin{flushleft}
\begin{tabularx}{\linewidth}{>{\raggedright\arraybackslash}X}
\toprule
\textbf{MCM Stage B}\\
\midrule
\textbf{Input:}\\
Mutated question and options \\
\textbf{Prompt:}\\
Choose the correct option and provide an explanation for each option.\\
\textbf{ChatGPT's response:}\\
Correct option: \textit{Y}\\
{Explanation for a: ...}\\
{Explanation for b: ...}\\
{Explanation for c: ...}\\
{Explanation for d: ...}\\
\textbf{Filtering:}\\
If X=Y, an instance for each option is created.\\
\bottomrule
\end{tabularx}
\caption{In Stage B of MCM, the teacher-LLM predicts the correct option again and generates explanations for each option.}
\label{tab:mcm_b}
\end{flushleft}
\end{minipage}

\end{table}

In the first prompting stage (see Table \ref{tab:mcm_a}), we provide an instance from the training data (introduction, question and answer) and ask GPT-3.5 to generate a multiple choice question with four options. Initially, the responses were often not satisfactory.
To improve them we introduced two prompt engineering artifacts, which we call `concept' and `question background', that clarify the type of questions we want to generate. These artifacts are demonstrated in the example provided with the prompt. The \mbox{`concept'} aims in identifying the legal outcome of the case and the `question background' aims in generating a different fictional scenario that is not related to the original one. You can find more details about these prompt engineering artifacts in Appendix \ref{sec:appendix_mcm_artifacts}.
The final prompt asks for a multiple choice question, with a different \mbox{`background'} than the original question, that illustrates the same \mbox{`concept'}. 
We then concatenate the `question background' and the `multiple choice question' that are both generated by the model to get the `mutated question' that we will use as synthetic data. We discard the `concept'. In the same prompt we also ask for the correct answer out of the candidate options.

For the second prompting stage (see Table \ref{tab:mcm_b}) we provide as input the output of the first prompting stage without the correct answer (concept, mutated question, candidate answers). We ask again GPT-3.5 to choose the correct choice and we also ask for an explanation that justifies each choice as correct or incorrect. To avoid introducing noisy synthetic training instances, if the option chosen in the first stage is not same as the option chosen in the second stage, meaning that GPT-3.5 answered inconsistently this question, the example is discarded. We call this process \textit{consistency filter}, as it is inspired from the self-consistency approach of \cite{wang2023selfconsistency}. Around 30\% of the generated examples were discarded. 

\section{Experimental setup}
\subsection{Data}
The initial data splits provided for the competition included a train split of 666 samples, a development (dev) split of 84 samples, and a test split of 98 samples. Out of the 84 samples of the development set, only 17 were labeled as correct. We found that these were not enough for our qualitative analysis (Section \ref{sec:qualitative}) and for this reason, we expanded the dev set by including 101 samples from the training set, resulting in 185 samples and a reduced training split of 565 samples. (We cannot conduct qualitative evaluation on the test set, as we do not possess the expert's analysis for those instances.)
The F1 score is the official evaluation measure due to the dataset's imbalance, with accuracy serving as the secondary metric.

\subsection{ChatGPT Prompting setup}
\label{section:LLMprompting}
All of our data augmentation was conducted with the model \texttt{gpt3.5-turbo-1106}. The total cost for data augmentation remained under \$20. We also use GPT models with few-shot prompting as baselines. For the experiments listed in Table \ref{main_results}, the scores on the development set were averaged over three different random seeds for GPT-3.5 and one seed for GPT-4.  In all experiments, the default OpenAI parameter values were kept.

\subsection{Llama-2-7B Tuning setup}
As the student model, we employed the 8-bit quantized version of the \texttt{Llama-2-7b} foundational  model.
We trained with QLoRA \cite{dettmers2023qlora} for one epoch, with a batch size of 4, and a learning rate of 1e-4. We utilized the Hugging Face Transformers library and the Llama recipes from Facebook Research.\footnote{\href{https://github.com/facebookresearch/llama-recipes}{github.com/facebookresearch/llama-recipes}}
 We used a single NVIDIA GPU A6000 with 48GB of GPU RAM.

\section{Results}

\subsection{Main Results}
Our best method ranked $15^\text{th}$ among 20 competitors. However, unlike other models, it can generate concise explanations to justify its answers. 
Our main baseline is a Llama-2 model (Llama-2-base) fine-tuned on the (reduced) training set without data augmentation.
Extending the dataset with human-guided explanations (Llama-2-HGE, Section \ref{hge}) greatly improves performance (Table \ref{main_results}) while providing insightful explanations as a byproduct. Employing additional synthetic data generated by the multiple choice mutation method (Llama-2-MCM, Section \ref{mcm}) along with the human-guided explanations further enhances performance. 
Additionally, Llama-2-MCM outperforms GPT-3.5 prompted with CoT examples (GPT-3.5-CoT), a strong baseline that can also provide explanations. This is a notable achievement for several reasons. First of all, GPT-3.5 produced the data that were used for fine-tuning Llama-2-MCM. Furthermore, it is a much more capable few-shot reasoner than Llama-2-7B \cite{leaderboard} and it is a significantly larger model in terms of parameters (probably more than ten times larger, but the exact number is unknown). 

\begin{table}[h]
\begin{tabularx}{\linewidth}{l*{4}{>{\centering\arraybackslash}X}} 
\toprule
\textbf{Model} & \textbf{F1 (dev)} & \textbf{Acc (dev)} & \textbf{F1 (test)} & \textbf{Acc (test)} \\
\midrule
BERT-base (F)& - & - & 56.80 &\textbf{80.22}\\
LegalBERT (F)& - & - & 65.73 & 76.92  \\
\midrule
GPT-3.5-base (P)& 31.92 & 32.97 & 27.60 & 30.61 \\
GPT-3.5-CoT (P) & 55.29 & 59.10 & 43.81 & 45.92\\
GPT-4-CoT (P)& \textbf{70.43} & \textbf{75.00} & \textbf{65.88} & 72.45 \\
\midrule
Llama-2-base (F)& 36.13 & 36.24 & 31.25 & 35.33 \\
Llama-2-HGE (F)& 50.88 & 65.08 & 50.00 & 59.18 \\
Llama-2-MCM (F)& 55.87 & 66.70 & 51.43 & 61.22 \\
\bottomrule
\end{tabularx}
\caption{Results for BERT-based models, prompted GPT models, and Llama fine-tuned models. (F) stands for fine-tuning and (P) stands for few-shot prompting. }
\label{main_results}
\end{table}

\subsection{BERT models}
We report BERT scores (Table \ref{main_results}) from the work that introduced the dataset \cite{legal_arg_reasoning_US_civil_procedure}. Even BERT-base without legal-specific pretraining achieves great performance, while LegalBERT with legal pretraining is exceptional, as it would rank $8^\text{th}$ in the competition and has similar results to a (prompted) GPT-4, a much larger (but not fine-tuned) model. However, as argued by \citet{legal_arg_reasoning_US_civil_procedure}, LegalBERT's responses are of limited utility, as they do not provide explanations of the reasoning behind them.

\subsection{OpenAI GPT models}
We evaluate GPT-3.5 and GPT-4, which are powerful general-purpose models, prompted with appropriate system instructions and one-shot examples. GPT-3.5-base, without CoT, acting only as a classifier model (no explanations) that predicts the correct label, performs poorly in both dev and test set (Table \ref{main_results}). Leveraging CoT (GPT-3.5-CoT) shows better performance, but still underperforms BERT-base and our Llama-2-MCM in the test set.  GPT-4-CoT outperforms GPT-3.5-CoT (by a large margin) and has similar performance with LegalBERT (would rank $8^\text{th}$ as well); in addition it can also produce explanations. Interestingly, the performance of GPT-3.5-CoT and GPT-4-CoT drops substantially in the test set if compared to the dev set. The decrease is larger in the case of GPT-3.5-CoT, probably because it is a less powerful model. Note that the test set is harder according to \citet{legal_arg_reasoning_US_civil_procedure}. 

The downsides of OpenAI models are that they are costly and cannot be fine-tuned in the specific domain as easily as open-source models.

\subsection{Llama models}
Our main baseline, Llama-2-base, is fine-tuned on the training set without data augmentation. It produces the predicted label only, without an explanation. In a similar fashion to its GPT-counterpart (GPT-3.5-base), it performs poorly, which highlights the importance of using CoT data either in a prompting or a fine-tuning setting. Llama-2-HGE (Section \ref{hge}), fine-tuned on the original data extended with explanations created by GPT-3.5 according to the expert's analysis, outperforms GPT-3.5-CoT on the test set.
Llama-2-MCM (Section \ref{mcm}), fine-tuned on HGE data along with artificial data generated by our `mutation' method, prevails over both Llama-HGE and GPT-3.5-CoT in every metric.
It falls short of BERT-base's performance. The boost in performance that further pretraining provides to LegalBERT might suggest that such an approach (domain adaptation) would benefit our model substantially as well.

\subsection{Ablation Study}
In the first rows of Table \ref{tab:ablation}, we report an experiment without the expert's analysis (no anls) to assess the effects of human guidance (Section \ref{hge}). Instead, we rely on CoT explanations from GPT-3.5's inherent knowledge, following \citet{mukherjee2023orca}.
The addition of the expert's analysis slightly improves performance, but for a more comprehensive assessment of the analysis's impact, manual evaluation of the explanations is necessary. However, we defer this aspect to future research. We also evaluate the impact of MCM's filtering process (Section \ref{mcm}). MCM's performance drops significantly (4 percentage points) without consistency filtering (no fltr, see Table \ref{tab:ablation}), highlighting its value in discarding poor synthetic examples. 

The plot in Fig.\hyperref[diagram]{ 1} shows the effect of using  more synthetic data from MCM. In the x-axis, we can see how many additional synthetic MCM instances are used for fine-tuning; 
the denominator is the number of synthetic data generated and the nominator (184, 344, 561, 709) are those kept after filtering. After all of the 565 original instances are utilized once, we add a second generation of synthetic data (red$ ^\triangle$ points) with a different random seed.
The second generation decreases performance, which may be attributed to the addition of two synthetic instances (that describe the same legal outcome) per original instance, potentially causing overfitting.

\begin{table}[h!]
\raisebox{\dimexpr\height}{ 
\begin{minipage}[t]{0.45\linewidth}
\small
\centering
\begin{tabularx}{\linewidth}{l*{2}{>{\centering\arraybackslash}X}} 
\toprule
\textbf{Model} & \textbf{F1 (dev)} & \textbf{Acc (dev)}\\
\midrule
No anls & 49.99 & 60.76 \\
HGE & 50.88 & 65.08 \\
\midrule
\midrule
No fltr& 51.14 & 69.41\\
MCM & 55.87 & 66.70 \\
\bottomrule
\end{tabularx}
\caption{The effects of the expert's analysis (anls) and the consistency filter (fltr).}
\label{tab:ablation}
\end{minipage}%
}
\hfill
\begin{minipage}[t]{0.5\linewidth}
\begin{flushleft}
\begin{tikzpicture}
\begin{axis}[
    width=\linewidth,
    xlabel={ $\frac{\small \text{ synthetic kept}}{\small \text{ all synthetic}}$  }, 
    ylabel={F1},
    ylabel style={rotate=-90, xshift=5pt, font=\small}, 
    xtick={0, 300, 565, 865, 1160},
    ytick={30, 35, 40, 45, 50, 51, 52, 53, 54, 55,56, 57, 58,60,65},
    xticklabels={0, $\frac{184}{300}$, $\frac{344}{565}$, $\frac{561}{865}$, $\frac{709}{1130}$},
    legend pos=outer north east,
    ymajorgrids=true,
    grid style=dashed,
    every tick label/.append style={font=\footnotesize}, 
    every axis label/.append style={font=\footnotesize} 
]

\addplot[color=red!80,mark=triangle*]
    coordinates {(565,55.87)(865, 52.51) (1160, 52.23)};
\addplot[color=blue!80,mark=otimes*]
    coordinates {(0, 50.88)(300,54.02)(565,55.87)};
\addplot[color=black,mark=square*]
    coordinates {(0, 50.88)};

\end{axis}
\end{tikzpicture}
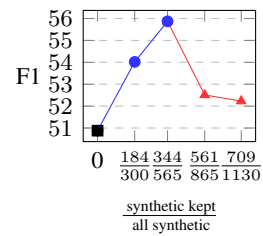
\caption*{Figure 1: Llama-2-MCM scores in dev set. More mutated examples from the same GPT-3.5 query (red$^\triangle$) do not improve F1 score.}
\label{diagram}
\end{flushleft}
\end{minipage}
\end{table}

\subsection{Qualitative Analysis}
\label{sec:qualitative}
We investigate how often the model fails to generate an accurate explanation, but nevertheless succeeds to predict the correct answer-label. For this we asked two legal experts (a professor in Law and a J.D. candidate) to annotate whether the explanations of 16 correct predictions of our Llama-2-MCM model align with the expert's analysis or not (Figure \hyperref[alignment]{ 2}). 8 of them were `aligned' and the other 8 were `not aligned'. This is an indication that performance can be deceiving, because a model might perform well without accurate reasoning.
In Table \ref{alignment_error} you can see an example that is `not aligned' due to a deficiency in legal knowledge.

The same experts were also asked to assess 16 false predictions of Llama-2-MCM in terms of `clarity', i.e., clear vs.\ unclear (see Fig.\hyperref[clarity]{ 3}) and comment on their observations. The objective of the second experiment was to investigate the nature of the model's errors.
12 out of 16 incorrect predictions were accompanied by clear explanations that enabled the experts to provide insightful feedback.
These explanations fall into two categories: reasoning deficiencies (4) and legal knowledge deficiencies (8).
Notably, the model demonstrated confusion regarding the distinction between federal and state courts in two instances, which is a critical aspect of legal knowledge.
Table \ref{incorrect_reasoning_example} presents an example in which the model appeared to comprehend the relevant legal knowledge, but ultimately failed due to constraints in its reasoning abilities.

\section{Conclusion}
This paper described the Archimedes-AUEB submission to SemEval-2024 Task 5. 
The submitted method is based on a LLM fine-tuned on CoT data in order to offer justifications for the predictions it generates. 
The training explanations are generated by GPT-3.5 in conjunction with expert analyses. 
Concurrently, we implement an innovative prompting procedure, dubbed Multiple Choice Mutation, designed to produce artificial instances.
Both approaches contribute to a significant increase in performance.
Our solution ranks $15^\text{th}$ out of $20$ contestants. Unlike other competitors, it generates explanations for its predictions. It also surpasses a strong GPT-3.5 baseline that can also produce explanations. 
We also assessed the output of our system qualitatively, with assistance from legal experts.
In many cases the system generated high-quality explanations.
The errors made by the model can be attributed to either its reasoning limitations or its deficiencies in legal knowledge.
We recommend including similar qualitative evaluations in future work, as they reveal pivotal insights about the model that performance metrics alone cannot capture.

\begin{figure}[h!]
    \centering
    \begin{minipage}[t]{0.45\linewidth}
        \centering
        \begin{tikzpicture}[scale=0.2]
        \pie[color={green!40,red!40}, sum=auto, text=legend, rotate=90]{8/\small{Aligned}, 8/\small{Not aligned}}
        \end{tikzpicture}
        \caption*{Figure 2: Correct predictions annotated for correct alignment with expert analysis.}
        \label{alignment}
        \vspace{5pt}
    \end{minipage}\hfill
    \begin{minipage}[t]{0.45\linewidth}
        \centering
        \begin{tikzpicture}[scale=0.2]
        \pie[color={cyan!40,brown!60}, sum=auto, text=legend, rotate=90]{12/\small{Clear}, 4/\small{Unclear}}
        \end{tikzpicture}
        \caption*{Figure 3: Incorrect predictions annotated for clarity.}
        \label{clarity}
    \end{minipage}
\end{figure}


\begin{table}[h!]
\small
\begin{adjustbox}{width=0.5\textwidth,center}
\begin{tabularx}{\linewidth}{>{\raggedright\arraybackslash}X}
\toprule
{\centering \textbf{Correct prediction but inaccurate explanation}}  \\
\midrule
\textbf{Question:}\\
A computer programmer living in Texas did some work for a corporation which is located in the Northern District of Illinois, but they refused to pay him.\\
\textbf{Correct answer:}\\
The court would have the authority to transfer to the Northern District of Illinois, but probably would not.\\
\textbf{Expert analysis:}\\
It is unlikely that the convenience of witnesses and the interests of justice require transfer of this case to Illinois.\\
\textbf{Model output:}\\
It would not be an appropriate venue, as the case should have been filed in the Texas district.\\
\bottomrule
\end{tabularx}
\end{adjustbox}
\caption{Example where the model predicted the correct label, but failed to produce an accurate explanation due to a deficiency in legal knowledge.}
\label{alignment_error}
\end{table}


\begin{table}[h!]
\small
\begin{adjustbox}{width=0.5\textwidth,center}
\begin{tabularx}{\linewidth}{>{\raggedright\arraybackslash}X}
\toprule
{\centering \textbf{Incorrect prediction but plausible explanation}}  \\
\midrule
\textbf{Question:}\\
A sues his former corporation, claiming that he was fired based on his age. B, one of his former supervisors, is about to move to South Korea. A’s counsel should\\
\textbf{Correct answer:}\\
prepare for the deposition, and cross-examine B as if he were doing so at trial.\\
\textbf{Model output:}\\
Incorrect, because a plaintiff who is planning to call B as a witness should not cross-examine her during the deposition to preserve her trial opportunities.\\
\textbf{Legal expert feedback:}\\
The AI failed to recognize that the witness planned to leave the country and there will not be any opportunity to examine him at trial.\\
\bottomrule
\end{tabularx}
\end{adjustbox}
\caption{Example where the model understood legal concepts, but failed due to reasoning limitations.}
\label{incorrect_reasoning_example}
\end{table}

\section*{Acknowledgements}
We are most grateful to the legal experts, 
Philippe Jougleux, Associate Professor, School of Law, European University Cyprus, and 
Helen Bougas, J.D. Candidate, Southern Methodist University.

This work has been partially supported by project MIS 5154714 of the National Recovery and Resilience Plan Greece 2.0 funded by the European Union under the Next Generation Program.

\bibliography{anthology,custom}

\appendix
\section{Appendix}
\subsection{Complete prompt examples}
\label{sec:appendix_prompts}
The following are examples with the complete prompts used for data augmentation with GPT-3.5 (see Table \ref{app:hge} for HGE, Table \ref{app:mcm_a} for MCM-A and Table \ref{app:mcm_b} for MCM-B). For each example we provide system instructions, an example of the task and the ideal response. During generation with GPT-3.5, we would also provide the actual training instance formatted as in the task example (Introduction, Question, Answer). To find the best prompts that worked well consistently we performed some initial experiments and inspected the quality of individual examples. We present examples of this process for Stage A of MCM in Appendix \ref{sec:appendix_mcm_artifacts}. 

\begin{table}[h!]
\small
\begin{adjustbox}{width=0.5\textwidth,center}
\begin{tabularx}{\linewidth}{>{\raggedright\arraybackslash}X}
\toprule
{\centering \textbf{Human-guided explanations}}  \\
\midrule
{\centering \textbf{System instructions}}  \\
You are an expert lawyer in the domain of U.S. civil procedure. You are given an introduction to a legal case, a question, an answer and an analysis that justifies the correctness of that answer. These are taken from a legal textbook, however they are not formal and they refer to multiple choice letters (A, B, C, D). Read the given analysis carefully and explain why the answer is correct in a short, formal, clear way, without mentioning anything about letters A, B, C, D.\\ 
{\centering \textbf{Task example}}  \\
Introduction:\\
Section 1391(d), which defines the residence of a corporation that has contacts in one district within a state but not others, is confusing. [...]\\
Question:\\Arthur wishes to bring a diversity action in federal court against Cleveland Manufacturing Company. Cleveland has its factory and principal place of business in the Northern District of Illinois, but no other contacts with Illinois. The claim is based on alleged negligence in making a toaster at the Illinois factory, which caused a fire in Arthur’s home in the Middle District of Georgia.\\
Answer:\\
The Southern District of Illinois is not a proper venue under §1391 because no events giving rise to the claim took place there and Cleveland does not reside there under the venue statute.\\
Analysis:\\
So, C is the right answer. The Southern District of Illinois is not a proper venue.\\
{\centering \textbf{Ideal response}}  \\
Explanation:\\
The Southern District of Illinois cannot serve as a proper venue for Arthur's lawsuit against Cleveland Manufacturing Company under Section 1391(d) because Cleveland lacks sufficient contacts in this district to be considered as 'residing' there within the venue statute.\\
\bottomrule
\end{tabularx}
\end{adjustbox}
\caption{Prompt instructions and one-shot example for generating human-guided explanations. The query for the training instance is not shown, but it follows the same structure as the `Task example'.}
\label{app:hge}
\end{table}

\begin{table}[h!]
\small
\begin{adjustbox}{width=0.5\textwidth,center}
\begin{tabularx}{\linewidth}{>{\raggedright\arraybackslash}X}
\toprule
\textbf{Multiple choice mutation - Stage A}\\
\midrule
{\centering \textbf{System instructions}}  \\
You are an expert lawyer in the domain of U.S. civil procedure. You are given an introduction that describes a legal case, a question about that case and an answer to that question. First, you must identify what legal concept the question is talking about. Note that the given question contains some background on the details of the case. Provide a different background for your question and then ask a multiple choice question in the style of the original question, to illustrate that concept to law students.\\
{\centering \textbf{Training instance}}  \\
Introduction:\\
... \\
Question:\\
Chu, a Californian, went skiing at Aspen, in Colorado, which has only one federal judicial district. After he discovered that he was somewhat out of shape, he bought an exercise machine from Jackson, a friend from Kansas with a two-week time share in Aspen, who had brought the machine with him to Aspen after buying it in Texas. The machine was made by Sweda- Trak, a Texas company that sells its products only in Texas, and with its only place of business in the Western District of Texas. Chu brought the machine back to Los Angeles, California, where he was injured using it. Chu brings a diversity action against Jackson and Sweda-Trak in federal court, alleging that Jackson was negligent in maintaining the machine, and Sweda-Trak in designing it. Venue in Chu’s action would be\\
Answer:\\
proper in the Southern District of California under §1391(b)(2), because a substantial part of the events giving rise to the claim took place there.\\
{\centering \textbf{Ideal response (Mutated instance)}}  \\
Concept:\\
Venue in Civil Procedure\\
Background:\\
Sophia, a New York resident, wants to sue Eric, a Texas resident, claiming that he infringed on her copyright by using her work in his recent publication. Sophia has never been to Texas, but she knows that Eric frequently visits New York for business meetings. According to U.S. law, venue refers to the most appropriate location for a trial. It can be the district where the defendant resides or where the events related to the claim occurred.\\
Question:\\
In which district would venue be proper if Sophia brings a copyright infringement lawsuit against Eric in federal court?\\
Choices:\\
A) The Southern District of New York, where Sophia resides\\
B) The Western District of Texas, where Eric resides\\
C) The Eastern District of New York, where Eric frequently visits for business meetings\\
D) The Northern District of Texas, where Sophia's work was originally created\\
Correct Answer:\\
A) The Southern District of New York, where Sophia resides\\
\bottomrule
\end{tabularx}
\end{adjustbox}
\caption{Prompt instructions and one-shot example for
generating a mutated question and candidate answers (choices/options) from an original training instance with MCM - Stage A.}
\label{app:mcm_a}
\end{table}

\begin{table}[h!]
\small
\begin{adjustbox}{width=0.5\textwidth,center}
\begin{tabularx}{\linewidth}{>{\raggedright\arraybackslash}X}
\toprule
\textbf{Multiple choice mutation - Stage B}\\
\midrule
{\centering \textbf{System instructions}}  \\
You are an expert lawyer in the domain of U.S. civil procedure. You are given an introduction that describes legal concepts, one of these concepts, background of a legal case and a multiple choice question about that case. First, answer which answer is the correct one and then for each choice explain in one sentence why it is correct or wrong. \\
{\centering \textbf{Task example}}  \\
Introduction:\\
...\\
Concept:\\Venue in Civil Procedure\\Background:\\Mark, a resident of Florida, wants to file a personal injury lawsuit against GreenCo, a multinational corporation headquartered in Delaware, for injuries sustained due to a defective product. The product was purchased online and delivered to Mark's home in Florida. Mark believes the defective product was manufactured and designed at GreenCo's facility in Kansas.\\Question:\\In which district would venue be proper if Mark brings a personal injury lawsuit against GreenCo in federal court?\\Choices:\\A) The District of Delaware, where GreenCo is headquartered\\B) The Middle District of Florida, where Mark resides\\C) The District of Kansas, where the defective product was manufactured and designed\\D) The Southern District of Florida, where the product was delivered to Mark's home\\ 
{\centering \textbf{Ideal response}}  \\
The correct answer is:\\B) The Middle District of Florida, where Mark resides\\Choices:\\A) The District of Delaware, where GreenCo is headquartered\\ Incorrect, because while GreenCo's headquarters are in Delaware, there doesn't seem to be any significant actions related to the claim that occurred there.\\B) The Middle District of Florida, where Mark resides\\ Correct, because the product was delivered to Mark's home in Florida, causing the injury. Therefore, a significant portion of the events related to the claim occurred there.\\C) The District of Kansas, where the defective product was manufactured and designed\\ Incorrect, although the product was manufactured and designed in Kansas, the significant harm occurred in Florida, making this district less relevant for this case.\\
D) The Southern District of Florida, where the product was delivered to Mark's home\\ Incorrect, because Mark does not reside in the Southern District of Florida and no specific information was given that suggest the Southern District as the proper venue.\\
\bottomrule
\end{tabularx}
\end{adjustbox}
\caption{Prompt instructions and one-shot example for
filtering inconsistent responses and generating explanations for each candidate answer from a mutated question and its candidate answers with MCM - Stage B.}
\label{app:mcm_b}
\end{table}

\clearpage

\subsection{Prompt engineering artifacts for MCM}
\label{sec:appendix_mcm_artifacts}
As explained in Section \ref{mcm}, the initial prompt we came up with was not generating consistent responses from GPT-3.5 and for this reason we added the artifacts of `concept' and `question background'. We experimented with many different prompts and inspected their outputs to optimize these artifacts. Here we present an example of a training instance in Table \ref{app:original_train} (it is the same example as in Table \ref{tab:correct-example}) and show the different mutated instances that would be produced with and without artifacts in Table \ref{app:prompt_with_artifacts} and Table \ref{app:naive_prompt} respectively. It is clear even from just one example that the response is much closer to the original training instance when we use artifacts. It is notable that we did not encounter the same problems with GPT-4.
\vfill
\begin{table}[h!]
\small
\begin{adjustbox}{width=0.5\textwidth,center}
\begin{tabularx}{\linewidth}{>{\raggedright\arraybackslash}X}
\toprule
{\centering \textbf{Original training instance}}  \\
\midrule
Introduction:\\
$\left[...\right]$ the Supreme Court addressed this issue in Klaxon Co. v. Stentor Mfg. Co., 313 U.S. 487 (1941). Klaxon held that a federal diversity court should use the choice-of-law rules of the state in which it sits.\\
...\\
Question:\\
Your law or mine? Whitney, a Rhode Island citizen, goes skiing at a ski area in Vermont. When he purchases his lift ticket, he quickly signs a release of liability, waiving his right to sue, should he be injured due to negligence of the operator. While coming down an expert trail, he comes suddenly upon a bare spot full of rocks. He falls and is injured. He brings a diversity action against the operator in federal court in Rhode Island, basing personal jurisdiction on the fact that it had sent him an offer there for a weekend ski package that induced him to go to Vermont to ski. Assume that Rhode Island tort law refuses to enforce advance waivers of liability for negligence, unless they were individually negotiated. Assume further that Vermont tort law holds such waivers enforceable. Assume further that Rhode Island’s choice-of-law rule for tort cases is to apply the law of the place of the injury, while Vermont’s choice-of-law rule involves an assessment of the interests of the parties and states involved. It is less predictable under this balancing approach to choice of law whether the court would apply Rhode Island or Vermont law. The Rhode Island federal district judge would probably apply\\
Answer:\\
Vermont law to Whitney’s case, because the accident happened in Vermont.\\
\bottomrule
\end{tabularx}
\end{adjustbox}
\caption{Original training instance example.}
\label{app:original_train}
\end{table}

\begin{table}[h!]
\small
\begin{adjustbox}{width=0.5\textwidth,center}
\begin{tabularx}{\linewidth}{>{\raggedright\arraybackslash}X}
\toprule
{\centering \textbf{Mutated question without artifacts}}  \\
\midrule
\textbf{Vanilla prompt}\\
\midrule
You are an expert lawyer in the domain of U.S. civil procedure. You are given an introduction that describes a legal case, a question about that case and an answer to that question. Ask a multiple choice question in the style of the original question.\\
\midrule
\textbf{Response}\\
\midrule
Question:\\
In a scenario where Emma, a Florida resident, sues Jack, a Georgia resident, for breach of contract in a federal court located in Georgia, which state's law would the federal court likely apply to the case under the principle established in Klaxon Co. v. Stentor Mfg. Co.?\\
Choices:\\
A) Florida law, where Emma resides\\
B) Georgia law, where Jack resides\\
C) Texas law, where the contract was signed\\
D) New York law, where the federal court is located\\
Correct Answer:\\
B) Georgia law, where Jack resides\\
\bottomrule
\end{tabularx}
\end{adjustbox}
\caption{The mutated question is missing some key elements when artifacts are not used, because GPT-3.5 has limited ability to understand complex prompts.}
\label{app:naive_prompt}
\end{table}
\begin{table}[h!]
\small
\begin{adjustbox}{width=0.5\textwidth,center}
\begin{tabularx}{\linewidth}{>{\raggedright\arraybackslash}X}
\toprule
{\centering \textbf{Mutated question with artifacts}}  \\
\midrule
\textbf{Prompt with artifacts}\\
\midrule
You are an expert lawyer in the domain of U.S. civil procedure. You are given an introduction that describes a legal case, a question about that case and an answer to that question. \textbf{First, you must identify what legal concept the question is talking about. Note that the given question contains some background on the details of the case. Provide a different background for your question} and then ask a multiple choice question in the style of the original question, \textbf{to illustrate that concept to law students.}\\
\midrule
\textbf{Response}\\
\midrule
Concept:\\
Choice of Law in Diversity Cases\\
Background:\\
Megan, a Florida resident, enters into a contract with Alex, a Georgia resident, for the purchase of a rare antique clock. The contract was negotiated over the phone and through emails, with the clock to be delivered to Megan in Florida. However, when the clock arrives, Megan discovers that it is damaged beyond repair. She decides to sue Alex in federal court in Georgia for breach of contract. The issue arises as to which state's law should govern the contract - Florida where Megan resides, or Georgia where Alex resides.\\
Question:\\
In a diversity action where Megan brings a breach of contract lawsuit against Alex, which state's law would a Georgia federal court likely apply, based on traditional choice-of-law doctrine?\\
Choices:\\
A) Florida law, where Megan resides \\
B) Georgia law, where Alex resides\\
C) The law of the state where the contract was formed, based on the traditional choice-of-law rule\\
D) The law of the state with the most significant contacts to the contract, based on the traditional choice-of-law rule\\
Correct Answer:\\
C) The law of the state where the contract was formed, based on the traditional choice-of-law rule\\
\bottomrule
\end{tabularx}
\end{adjustbox}
\caption{The mutated question is closer to our desired format when we use artifacts.}
\label{app:prompt_with_artifacts}
\end{table}
\vfill
\clearpage

\subsection{System response samples}
\label{sec:appendix_responses}

In this section we present samples of complete instance-response pairs
from our system (from the extended development set) annotated by legal experts. We present two samples of alignment between expert analysis and model explanation (Tables \ref{app:alignment-1} and \ref{app:alignment-2}) and one sample that the explanation was not aligned (Table \ref{app:mis-alignment}). We also present one sample where the prediction was wrong and the explanation was inacurrate because of legal deficiencies (Table \ref{app:clear-federal}), another one that failed because of limitations in reasoning cabilities (Table \ref{app:clear-reasoning}) and a sample that failed but the explanation was unclear to the legal experts (Table \ref{app:unclear}).

\begin{table}[h!]
\small
\begin{adjustbox}{width=0.5\textwidth,center}
\begin{tabularx}{\linewidth}{>{\raggedright\arraybackslash}X}
\toprule
{\centering \textbf{Aligned explanation - 1}}  \\
\midrule
\textbf{Introduction:}\\
The message of the Erie case is clear: ‘‘federal court: apply state law in a diversity case.’’ 
$[...]$ 
Every state has ‘‘choice-of-law’’ rules to deal with this problem, which arises frequently in state cases as well as federal diversity cases.
$[...]$ 
Three years after Erie, the Supreme Court addressed this issue in Klaxon Co. v. Stentor Mfg. Co., 313 U.S. 487 (1941). Klaxon held that a federal diversity court should use the choice-of-law rules of the state in which it sits. Thus, a New York federal court should apply whatever state substantive law the New York state court would apply to the case. If a New York court, under its choice-of-law rules, would apply New York tort law to Tompkins’s accident (even though it took place in Pennsylvania), the New York federal court should do so as well. On the other hand, if a New York court would, under its choice-of-law rule for torts, apply Pennsylvania law to Tompkins’s case, the New York federal court should similarly look to Pennsylvania law.\\
...\\
\textbf{Question:}\\
Your law or mine? Whitney, a Rhode Island citizen, goes skiing at a ski area in Vermont. When he purchases his lift ticket, he quickly signs a release of liability, waiving his right to sue, should he be injured due to negligence of the operator. While coming down an expert trail, he comes suddenly upon a bare spot full of rocks. He falls and is injured. He brings a diversity action against the operator in federal court in Rhode Island, basing personal jurisdiction on the fact that it had sent him an offer there for a weekend ski package that induced him to go to Vermont to ski. Assume that Rhode Island tort law refuses to enforce advance waivers of liability for negligence, unless they were individually negotiated. Assume further that Vermont tort law holds such waivers enforceable. Assume further that Rhode Island’s choice-of-law rule for tort cases is to apply the law of the place of the injury, while Vermont’s choice-of-law rule involves an assessment of the interests of the parties and states involved. It is less predictable under this balancing approach to choice of law whether the court would apply Rhode Island or Vermont law. The Rhode Island federal district judge would probably apply\\
\textbf{Answer:}
Vermont law to Whitney’s case, because the accident happened in Vermont.\\
\textbf{Expert analysis:}\\
Label: \textit{Correct}\\
Although Rhode Island wouldn’t enforce this waiver in a domestic case, its choice-of-law rule calls for application of the law of the place of injury, and that means Vermont law rather than its own.\\
\textbf{Model explanation:}\\
Label: \textit{Correct}\\
The Rhode Island federal district judge would probably apply Vermont law to Whitney's case, as per Klaxon holding that the judge should use the local state's choice-of-law rules, resulting in Vermont law being likely applied.\\
\bottomrule
\end{tabularx}
\end{adjustbox}
\caption{An example of alignment between expert analysis and model explanation (this is the complete prompt and response for the instance that was shown in Table \ref{tab:correct-example}).}
\label{app:alignment-1}
\end{table}


\begin{table}[h!]
\small
\begin{adjustbox}{width=0.5\textwidth,center}
\begin{tabularx}{\linewidth}{>{\raggedright\arraybackslash}X}
\toprule
{\centering \textbf{Aligned explanation - 2}}  \\
\midrule
\textbf{Introduction:}\\
Rule 8 contains simple, sensible provisions governing the defendant’s responses to the allegations in the complaint. First, Rule 8(b) requires the defendant to admit or deny each of the allegations in the complaint.\\
$[...]$\\
Under Rule 8(b)(2), a ‘‘denial must fairly respond to the substance of the allegation.’’ If the defendant means to deny only a part of an allegation, she must specify what is true and deny only the parts that are not. Failing to deny an allegation constitutes an admission of it. Fed. R. Civ. P. 8(b)(6). The idea is that, if the parties plead conscientiously by the Rules, the parties and the court can compare the allegations in the complaint and the responses in the answer, and determine which issues are contested and which are agreed upon (and therefore need not be litigated).\\
\textbf{Question:}\\
Patterson sues Applied Electronics, Inc., for injuries suffered in an accident with Adair, one of AE’s delivery drivers. Patterson alleges in Paragraph 4 of his complaint that Adair was an employee of AE at the time of the accident, that Adair was acting in the scope of employment at the time of the accident (which Patterson must prove in order to recover from the employer), that Adair was talking on his cell phone at the time of the accident, and that Adair’s negligent driving caused the accident. AE answers Paragraph 4 as follows: ‘‘The defendant admits that Adair was its employee at the time of the accident, denies that Adair was negligent, but is without sufficient information to form a belief as to the truth of the allegation that Adair was talking on his cell phone at the time of the accident.’’ Under Federal Rule 8,\\
\textbf{Answer:}\\
AE has admitted that Adair was its employee, but not that he acted in the scope of employment. The allegation that Adair was talking on his cell phone is treated as denied.
\textbf{Expert analysis:}\\
Label: \textit{Incorrect}\\
This question simply requires you to compare Patterson’s allegations with the defendant’s responses, and apply the provisions of Rule 8 as to the effect of a denial, failure to deny, or allegation that the pleader is ‘‘without sufficient facts to form a belief.’’  Naturally, AE has admitted that Adair was its employee, by admitting it! (Sometimes even the pleading rules are commonsensical.) The allegation that Adair was talking on his cell phone is treated as denied: Under the Rule 8(b)(5), the ‘‘without sufficient information’’ response is treated as a denial. Adair hasn’t responded at all to the allegation that Adair acted in the scope of employment at the time of the accident. Frequently, defendants would like to ignore allegations in the complaint, if they aren’t sure of their truth, or it would be awkward to admit them. But this head-in-the-sand strategy doesn’t work under the Rules: Rule 8(b)(6) provides that allegations that are not denied are deemed admitted. So, effectively, AE has admitted that Adair was its employee and that he acted in the scope of employment, but denied negligence and that Adair was chatting on the cell phone.\\
\textbf{Model explanation:}\\
Label: \textit{Incorrect}\\
The answer is incorrect because the language used by AE is not consistent with the treatment described in Rule 8(b)(6), which requires the defendant to deny only a part of the accusation if not admitting the whole.\\
\bottomrule
\end{tabularx}
\end{adjustbox}
\caption{Another example of alignment between expert analysis and model explanation. Note that the model's explanation is much more concise.}
\label{app:alignment-2}
\end{table}

\begin{table}[h!]
\small
\begin{adjustbox}{width=0.5\textwidth,center}
\begin{tabularx}{\linewidth}{>{\raggedright\arraybackslash}X}
\toprule
{\centering \textbf{Failed alignment of explanation}}  \\
\midrule
\textbf{Introduction:}\\
$[...]$\\
In response to the Gulf Oil decision, Congress enacted 28 U.S.C. §1404(a), the federal transfer-of-venue statute. Under this section, a judge may transfer a case filed in her court that should be litigated elsewhere to the more appropriate federal district, instead of dismissing it for forum non conveniens. (a) For the convenience of parties and witnesses, in the interest of justice, a district court may transfer any civil action to any other district or division where it might have been brought or to any district or division to which all parties have consented. This statute makes good administrative sense. Because the federal district courts are all units of the same system, Congress has the authority to provide for transfer of cases among them to promote efficiency.\\
...\\
\textbf{Question:}\\
The federal shuffle. Connors, a computer programmer living in the Northern District of Texas, did some contract work for Ace Corporation, which is incorporated in Delaware with its principal place of business in the Northern District of Illinois. Ace has an office in the Northern District of Texas, but no other offices in Texas. It has large operations in three other states. Connors negotiated the contract by phone and e-mail from his home, dealing with Ace employees in the Texas office and the Illinois office. He did the programming work, which was to be used for Ace’s Texas operation, at his home. Ace Corporation claimed that Connors had only done half the work, and refused to pay. Connors, believing that he had completed all the work called for under the contract, brought a diversity action against Ace for the contract price (\$200,000) in federal court in the Northern District of Texas.\\
\textbf{Answer:}\\
The court would have the authority to transfer to the Northern District of Illinois, but probably would not.\\
\textbf{Expert analysis:}\\
Label: \textit{Correct}\\
Although the court would have authority under 28 U.S.C. §1404(a) to transfer this case to the Northern District of Illinois, why would it? Most of the events giving rise to the claim took place in the Northern District of Texas. Connors did the work there, dealt with Ace employees there, and suffered the damages there. The programming was to be used in Ace’s Texas office. The defendant, a corporation with a local office, would not be greatly inconvenienced or put to great expense to litigate this case in Texas. Some witnesses and evidence may be in Illinois, but it seems very doubtful that more would be there than in Texas. It seems highly unlikely that the convenience of witnesses and the interests of justice require transfer of this case to Illinois.\\
\textbf{Model explanation:}\\
Label: \textit{Correct}\\
The federal court in Texas would have the authority to transfer the case to the Northern District of Illinois under §1404(a), but the court would probably not do so because it would not be an appropriate venue, as the case should have been filed in the Texas district.\\
\bottomrule
\end{tabularx}
\end{adjustbox}
\caption{A sample that the model explanation failed to align with the expert's analysis (this is the complete prompt and response for the instance that was shown in Table \ref{alignment_error}).}
\label{app:mis-alignment}
\end{table}


\begin{table}[h!]
\small
\begin{adjustbox}{width=0.5\textwidth,center}
\begin{tabularx}{\linewidth}{>{\raggedright\arraybackslash}X}
\toprule
{\centering \textbf{Incorrect response due to legal deficiency}}  \\
\midrule
\textbf{Introduction:}\\
In Hanna, the Court addressed the power of Congress to promulgate rules that govern procedure in the federal courts. See 380 U.S. at 464. Analytically, it had to do so, because the power to write the Federal Rules of Civil Procedure is granted by Congress to the Court, in the Rules Enabling Act. Surely, Congress could not delegate to the Court any broader rule-making power than it possesses itself. In Hanna, the Court describes Congress’s rule-making authority as follows: the constitutional provision for a federal court system (augmented by the Necessary and Proper Clause) carries with it congressional power to make rules governing the practice and pleading in those courts, which in turn includes a power to regulate matters which, though falling within the uncertain area between substance and procedure, are rationally capable of classification as either. 380 U.S. at 472. That’s a mighty broad test for Congressional power over federal court procedure: If the provision can rationally be viewed as regulating procedure, Congress may enact it, even if it will affect the outcome of litigation. Consider, however, how the Court would rule if Congress enacted a statute that impinges on state law rights that can reasonably be classified as ‘‘substantive.’’\\
\textbf{Question:}\\
Limits on limitations. Congress, in an effort to avoid complex and time-consuming litigation on the proper time for filing diversity cases in the federal courts, enacts the Federal Diversity Limitations Act (FDLA), establishing a two-year limitations period for diversity cases in the federal courts, regardless of the applicable state limitations period for the claim. Sanchez brings a diversity action against Bernstein in federal court for fraud arising out of a real estate transaction. The relevant state limitations period for fraud claims is four years. Sanchez’s suit is brought two years and five months after the alleged fraud was committed. The federal court should\\
\textbf{Answer:}\\
dismiss the claim, because the FDLA applies and bars it.\\
\textbf{Expert analysis:}\\
While this limitations period would not apply if it were a matter of federal judicial practice (see York), and probably could not be validly adopted as a Federal Rule (see 28 U.S.C. §2072(b)), there’s a strong argument, under Hanna’s description of congressional power over procedure, that this statute could be enacted by Congress. After all, under Hanna, Congress has very broad authority to regulate procedure in the federal courts. You could imagine that Congress might seek to protect those courts from expending limited judicial resources on state claims by imposing a limitations period on diversity cases. It is very likely within its power to do so, and should be. Of course, it hasn’t, and probably won’t pass such a limitations statute, since such a statute would interfere significantly with the administration of state causes of action. The members of Congress do, after all, represent the states.\\
\textbf{Model explanation:}\\
Label: \textit{Incorrect}\\
The answer is incorrect because federal courts should not dismiss a case on the basis of federal statutes of limitations if state law permits the claim to be raised, unless there are grounds not based on the applicable federal limitations period to question admissibility of the claim.\\
\textbf{Legal expert feedback:}\\
The system has difficulty distinguishing the differences between state and federal court.\\
\bottomrule
\end{tabularx}
\end{adjustbox}
\caption{An example where the model demonstrated confusion in distinguishing the state with the federal court.}
\label{app:clear-federal}
\end{table}


\begin{table}[h!]
\small
\begin{adjustbox}{width=0.5\textwidth,center}
\begin{tabularx}{\linewidth}{>{\raggedright\arraybackslash}X}
\toprule
{\centering \textbf{Incorrect response due to limited reasoning capabilities}}  \\
\midrule
\textbf{Introduction:}\\
A deposition is the examination of a witness under oath. Depositions are the most effective means of previewing the detailed testimony of witnesses.\\
$[...]$\\
Depositions serve several purposes. They ‘‘get the witness on the record.’’ That is, the witness tells her story under oath. If she contradicts that story at trial, she can be impeached with her deposition testimony.\\
$[...]$\\
However, in some cases a party may depose a witness in order to obtain the party’s testimony for use at trial. If, for example, a witness is moving out of the country, the party who wants to use that witness’s testimony at trial may take her deposition as a substitute for her live testimony at the trial. The deposition will be admissible at trial if the witness is not available to testify in person. Fed. R. Civ. P. 32(a)(4). Note that in this scenario, unlike the typical deposition, the party noticing the deposition will be the party who plans to use that witness’s testimony, not the opposing party.\\
\textbf{Question:}\\
Your witness.\\
Czonka is fired from his job at fifty-nine, after twenty-one years with NorthStar Corporation. He sues in federal court under the Federal Age Discrimination in Employment Act (ADEA), claiming that he was fired based on his age. Two months after filing suit, Czonka’s counsel learns that Roh, one of his supervisors at NorthStar, has just retired, and is about to move to South Korea. Although discovery has just begun, NorthStar has noticed Roh’s deposition. Czonka’s counsel should\\
\textbf{Answer:}\\
prepare as fully as possible for the deposition, and cross-examine Roh as if she were doing so at trial.\\
\textbf{Expert analysis:}\\
Label: \textit{Correct}\\
If Roh does not appear at trial, cross-examining at the deposition will be the only opportunity Czonka’s counsel has to cross- examine Roh under oath. She had better do it as well as she can.\\
\textbf{Model explanation:}\\
Label: \textit{Incorrect}\\
The answer is incorrect because a well-prepared plaintiff who is planning to call Roh as a witness should not cross-examine her during the deposition to preserve her trial opportunities, but should focus on clarifying points and getting her story on the record.\\
\textbf{Legal expert feedback:}\\
AI based the reasoning on trial opportunity whereas the protagonist planned to leave the country and there will not be any opportunity to examine him at trial.\\
\bottomrule
\end{tabularx}
\end{adjustbox}
\caption{An example where the model failed, even though it understood legal knowledge, due to limited reasoning capabilities (this is the complete prompt and response for the instance that was shown in Table \ref{incorrect_reasoning_example}).}
\label{app:clear-reasoning}
\end{table}

\begin{table}[h!]
\small
\begin{adjustbox}{width=0.5\textwidth,center}
\begin{tabularx}{\linewidth}{>{\raggedright\arraybackslash}X}
\toprule
{\centering \textbf{Incorrect response with unclear explanation}}  \\
\midrule
\textbf{Introduction:}\\
The discussion so far has focused on motions by the defendant. But the plaintiff may seek summary judgment as well, under the same standard: that there is no genuine issue of material fact and she is entitled to judgment as a matter of law. However, the effect of summary judgment for the plaintiff is often different from the grant of the motion for a defendant.\\ 
\textbf{Question:}\\
Insulation from liability. Maxwell sues Chao for breach of a contract to insulate Chao’s house. He alleges that they had a contract, and that he did the work, but Chao refused to pay the agreed price for the work. Maxwell seeks recovery in the alternative on a quantum meruit theory. Chao denies that Maxwell can sue for breach of contract, because the writing Maxwell relies on as the contract—a scribbled note indicating the price for the job—fails to satisfy the requirements for a valid contract under the relevant statute of frauds. Chao also claims that the insulation used was substandard and improperly installed. Maxwell moves for summary judgment on the question of whether the parties had a contract. In support of his motion, he submits the scribbled note itself, an affidavit attesting that it was signed by Chao in his presence, and a brief arguing that the note is sufficient to constitute a binding contract under the statute of frauds. Chao submits no opposing materials, but submits a brief arguing that the note does not constitute a contract, because the terms are insufficiently described in it.
\textbf{Answer:}\\
If the judge concludes that the note meets the requirements of a valid contract, she should enter partial summary judgment for Maxwell on the issue of the validity of the written contract.\\
\textbf{Expert analysis:}\\
Label: \textit{Correct}\\
If the judge concludes that the note suffices as a writing, she should grant partial summary judgment for Maxwell on the issue of the sufficiency of the note as a contract. But she would not enter judgment for Maxwell on the claim as a whole, because his motion only establishes one element of the claim, existence of the contract. Maxwell still has to establish other elements of the contract claim (such as adequate performance) or the quantum meruit claim, in order to recover from Chao.\\
\textbf{Model explanation:}\\
Label: \textit{Incorrect}\\
The answer is incorrect because while a judge may enter partial summary judgment on the issue of the written contract's validity, the plaintiff's alternative claim is the subject of different issues, and partial summary judgment will only cover those issues where there is no genuine issue of material fact.\\
\bottomrule
\end{tabularx}
\end{adjustbox}
\caption{An example of an incorrect explanation that the experts found unclear.}
\label{app:unclear}
\end{table}

\clearpage
\subsection{Experts' evaluation}
\label{sec:appendix_experts}
For the manual evaluation process by legal experts we created Google Forms automatically with the Google Apps Script software.\footnote{\url{https://www.google.com/script/start/}} Each form contains eight examples in total, four of which are correct labels and the other four incorrect. The experts were given general instructions (Table \ref{app:gen-instructions}) in order to fill in the forms as consistently as possible. They were also given specific instructions for each task (Table \ref{app:cor-instructions}). Each example in the form contained an input instance(introduction, question, answer and label of answer, expert analysis) and the prediction of Llama-2-MCM along with the generated explanation for that particular instance.

\begin{table}[h!]
\small
\begin{adjustbox}{width=0.5\textwidth,center}
\begin{tabularx}{\linewidth}{>{\raggedright\arraybackslash}X}
\toprule
{\centering \textbf{General Instructions for experts}}  \\
\midrule
\textbf{Outlined Steps}\\
Carefully review the Question, Answer, and corresponding Labels.
Read the Human Explanation briefly to gain a preliminary understanding.
Read the AI explanation and evaluation question carefully.
If you know the evaluation question, provide a direct response. Otherwise, revisit the skipped sections for further clarification.
In cases of uncertainty after revisiting the essential sections, you can proceed to the subsequent question. The objective is to provide definitive answers wherever possible while optimizing time efficiency.
Note that sometimes the explanations (either human or AI), refer to letters, such as "A is correct". Please ignore these, as the letters used to refer to multiple choice questions, but they have been shuffled.\\
Disclaimer: All legal scenarios presented are fictitious and derived from 'The Glannon Guide To Civil Procedure: Learning Civil Procedure Through Multiple-Choice Questions and Analysis, 4th edition.'\\
\bottomrule
\end{tabularx}
\end{adjustbox}
\caption{General instructions provided to legal experts before they fill in the Google Form.}
\label{app:gen-instructions}
\end{table}

\begin{table}[h!]
\small
\begin{adjustbox}{width=0.5\textwidth,center}
\begin{tabularx}{\linewidth}{>{\raggedright\arraybackslash}X}
\toprule
\textbf{Correct Predictions}\\
\midrule
In this scenario, the AI has correctly predicted an outcome, and we aim to assess whether it did so for the right reasons or merely by chance. To accomplish this, carefully read both the Question and the Answer, followed by the Human explanation. Next, review the AI explanation and determine if it seems plausible and consistent with the human explanation. Consider whether the AI truly comprehended the reasoning behind the answer. Please refrain from comparing it to the Human explanation, as our intention is not to ascertain whether the AI outperforms the human (which it does not, of course). If you're unsure about what the AI actually understood, you may leave your answer blank, but we encourage you to respond to as many samples as possible.\\
\midrule
\textbf{Incorrect Predictions}\\
\midrule
In this scenario, the AI has provided an incorrect prediction, resulting in an erroneous explanation. However, our objective is to assess the frequency with which the AI engages in what is known as "hallucination", where it imagines an explanation without genuine reasoning, versus instances where it genuinely attempts to reason but arrives at an incorrect conclusion. In the former case, akin to a student failing to put effort into their homework, we aim to identify instances where the AI requires correction. If you find the purpose of the AI explanation unclear, please indicate as such.\\
\bottomrule
\end{tabularx}
\end{adjustbox}
\caption{Instructions for the alignment task (Correct) and the task about the clarity of incorrect predictions (Incorrect) provided to legal experts before they fill in the Google Form.}
\label{app:cor-instructions}
\end{table}

\end{document}